# Putting GPT-3's Creativity to the (Alternative Uses) Test

Claire Stevenson, Iris Smal, Matthijs Baas, Raoul Grasman & Han van der Maas
Psychological Methods Department
University of Amsterdam
Roetersstraat 11, 1018 WB Amsterdam, The Netherlands
c.e.stevenson@uva.nl

## Abstract

AI large language models have (co-)produced amazing written works from newspaper articles to novels and poetry. These works meet the standards of the standard definition of creativity: being original and useful, and sometimes even the additional element of surprise. But can a large language model designed to predict the next text fragment provide creative, out-of-the-box, responses that still solve the problem at hand? We put Open AI's generative natural language model, GPT-3, to the test. Can it provide creative solutions to one of the most commonly used tests in creativity research? We assessed GPT-3's creativity on Guilford's Alternative Uses Test (AUT) and compared its performance to previously collected human responses on expert ratings of originality, usefulness and surprise of responses, flexibility of each set of ideas as well as an automated method to measure creativity based on the semantic distance between a response and the AUT object in question. Our results show that -on the whole- humans currently outperform GPT-3 when it comes to creative output. But, we believe it is only a matter of time before GPT-3 catches up on this particular task. We discuss what this work reveals about human and AI creativity, creativity testing and our definition of creativity.

## Introduction

A New York Times magazine headline (April, 2022) states "OpenAI's GPT-3 and other neural nets can now write original prose with mind-boggling fluency…". Reactions to this on Twitter and in blogposts vary, but many converge on the sobering belief that large language models (LLMs) are somewhat of a hype and err on the side of 'stochastic parrots', a reference to the computational linguists Bender et al. (2021) comments on the uses and dangers of large language models. We could easily take this a stretch further and argue LLMs have not achieved general artificial intelligence (Mitchell, 2021; van der Maas, Snoek and Stevenson, 2021), much less "truly" creative artificial creativity.

In daily life, such LLMs, and GPT-3 in particular, have proved very useful in (co-)creating phenomenal works: newspaper articles (e.g., GPT-3, 2020), novels (e.g., Green, 2020), and poetry (e.g., Aalho, 2021). The first author even has students who have admitted to using LLMs to help write their theses. These works meet the criteria of the standard definition of creativity: being original and useful, and sometimes even the additional element of surprise (Runco and Jaeger, 2012). But, how much creative thinking can be attributed to such LLMs? Can such large language models really produce the creative insights that humans are capable of?

In this study we put one particular LLM's creativity to the test, OpenAI's GPT-3 (Brown et al., 2020). We compare its performance to that of humans on the popular Alternative Uses Test (AUT, Guilford, 1967). On the AUT people are asked to produce as many original uses for an everyday object as possible (e.g., a 'brick' can be used as a 'paperweight' or 'to break a window'). Responses to the AUT are generally scored in terms of quality, such as the originality and utility of each generated idea, often rated by two or more experts using the Consensual Assessment Technique (Baer and McKool, 2009). In this study, we examine these two quality dimensions, where there is generally a trade-off between originality and utility (Rietzchel, Nijstad and Stroebe, 2019, as well as the surprise elicited by AUT responses as suggested by Boden (2004) and Simonton (2018). Surprise, where a response violates expectations and elicits interest, may be of particular interest when examining AI creativity in the context of LLMs. Given that LLMs are trained on nearly all the text on the Internet to -in essence- predict a text fragment given only the surrounding text, it would seem difficult for an LLM to generate a surprising, out-of-context response to the creative uses question.

A more recent method of gauging how creative a response is, is to measure the semantic distance of the response from the AUT object, a process automated with the SemDis software (Beaty and Johnson, 2021). SemDis measures are related to expert ratings of creativity and can be used as a proxy for creativity scoring (Beaty and Johnson, 2021).

Another method of analyzing a set of responses is focused more on the response process (Nijstad et al., 2010). Do most responses come from one conceptual space (e.g., using a brick as a paperweight, doorstop and bookend; so, to hold things in place)? Or is the response pattern more flexible, where numerous conceptual spaces are traversed (e.g., using a brick as a paperweight, sidewalk chalk and hot water bottle)? Hass (2017) found that whereas people often respond in clusters on semantic fluency tasks, such as listing as many animals as they can within a minute (e.g., naming pets, then naming farm animals, then zoo animals, etc.), people tend to use a more flexible strategy on the AUT. With GPT-3 being a predictive model, will it show a similar flexible response pattern?

To our knowledge this study represents the first systematic psychological assessment of a LLM's creativity. We compare how humans and GPT-3 score in terms of expert ratings of originality, usefulness and surprise of responses, automated semantic distance scoring as a proxy for creativity, and more holistically examine the flexibility of responses within a response set.

## Methods

### Sample

The human sample comprised of previously collected data of 823 responses from 42 students from the University of Amsterdam. Only data from students fluent in Dutch, the language of the study, were invited to participate. Written informed consent for participation was obtained and participants received study credits for participation. The data collection and subsequent re-use of the data was approved by our Psychology Department's Ethical Review Board (ERB number 6990).

The GPT-3 sample comprised of 144 runs of the AUT using the instructions and parameter settings described under Materials and Procedure.

### Materials

**Alternative Uses Task for humans**

We used a computerized version of the Alternative Uses Test (AUT; Guilford, 1967). Participants were given the name of an object and instructed to "Think of as many creative uses for" the object as possible within a two minute period. In this study we use the data from the "Book", "Fork", and "Tin Can" objects. Participants were instructed to "Type each solution in the text box below and press Enter to add it to the list.". The solutions remained on the screen until the time limit was reached.

**Alternative Uses Task for GPT-3**

We used Open AI's API to request responses from GPT-3 for each of the same objects administered to humans: "Book", "Fork", and "Tin Can".

Before collecting GPT-3's AUT data for this study, we conducted two Monte Carlo experiments to determine: (1) which GPT-3 engine performed best on the AUT and (2) which prompt and parameter settings let to the most valid responses (i.e. responses that answered the question, did not contain nonsense) and provided the highest snapshot creativity scores (Silvia et. al., 2008). See osf.io/vmk3c/ for the code, data and results of our optimization studies.

Based on these results we administered the AUT to GPT-3's davinci-002 engine as follows. The instruction was: "What are some creative uses for a [book|fork|tin can]? The goal is to come up with creative ideas, which are ideas that strike people as clever, unusual, interesting, uncommon, humorous, innovative, or different. List [9|10] creative uses for a [book|fork|tin can]." The most important parameter settings that differed from the default were the temperature (sampled from range .65 - .80), the frequency penalty (set to 1), and the presence penalty (also set to 1). We collected 820 responses from GPT-3 over two sessions.

### Procedure

Before we could score the responses we needed to make sure the judges could not easily distinguish between human and GPT-3 responses. First, we translated the 823 Dutch language human responses to English so that all responses were in the same language. Second, we removed characteristic punctuation from the GPT-3 data (e.g., numbered responses, period at the end of each line). Third, we systematically removed common phrases such as "Use a {object} to", "A {object} can be used to make", which is a step we usually take to make the rating of responses easier for human judges, which also happened to occur more often in the GPT-3 data.

After ensuring that GPT-3 and human responses were indistinguishable in all regards except main content, two trained judges rated each response on originality, utility, and surprise using a 5-point scale (from 1 = "not original | useful | surprising" to 5 = "highly original | useful | surprising") according to pre-specified scoring protocols. Judges were blinded to whether or not the responses stemmed from humans or GPT-3. The inter-rater agreement (assessed for approximately 10% of the 1656 responses) was ICC=.57 for originality, .68 for utility and .67 for surprise, which is considered fair to good. After removing invalid and incomplete responses (which were also excluded from analyses), the ICC's were .78 for originality, .70 for utility and .79 for surprise, which is considered good to excellent.

We computed the semantic distance (i.e., 1 – cosine similarity) between the vector embeddings for each response and the AUT object in question, which serves as a proxy for creativity (Beaty and Johnson, 2021), using the spaCy library and en_core_web_lg semantic space.

We also computed the flexibility of a response set. This was done by categorizing each response into one or more pre-defined categories and then dividing the number of categories in a response set by the number of responses in total in the response set. For example, if five responses were given to the AUT fork and each of these responses fell under the categories of "utensil" or "make music", then the resulting flexibility score would be 2 categories / 5 responses = 0.4 Three trained judges categorized the responses (one judge per object), by assigning each response to one or more pre-defined categories.

All data, code, Dutch to English translations, data cleaning steps, rating protocols, and categorization protocols are available on the Open Science Foundation website for this project: http://osf.io/vmk3c/.

## Results

In total, after data cleaning, we analyzed 774 responses from 42 humans and 781 responses from 144 GPT-3 runs. All data and analysis code can be found on http://osf.io/vmk3c/.

| | |
|---|---|
| 1. Use a tin can as a mirror | 1. Plant a herb garden in tin cans |
| 2. to create toys | |
| 3. to create jewelry | 2. Make a wind chime out of tin cans and beads |
| 4. as wallpaper as long as you stick enough next to each other | 3. as candle holders for an outdoor party |
| 5. to throw | |
| 6. as a knife | 4. Create a mini Zen garden in a tin can |
| 7. as a key ring with the clip from the can | 5. Make a robot out of recycled materials, including tin cans |
| 8. As a rattle with the clip in tin | |
| 9. as art | |
| 10. As a reminder of Andy Warhol | 6. Turn a tin can into a night light |

Figure 1. Two sets of responses to the AUT "Tin Can" task. Can you guess which one was given by a human and which one by GPT-3?[1]

**Do humans or GPT-3 provide more original, useful or surprising responses?**

We used hierarchical regression models to predict originality, utility and surprise ratings and the semantic distance scores at the response level, while accounting for correlations between responses within a person or a GPT-3 run. The predictors were a human versus GPT-3 contrast plus the AUT object (3 levels).

As can be seen in Figures 2, 3 and 4, our results showed that humans had higher originality ($\beta$=.17, SE = .06, $z$ = 2.91, $p$ = .004) and surprise ($\beta$=.14, SE = .07, $z$ = 1.96, $p$ = .050) ratings as well as larger semantic distance scores ($\beta$= .10, SE = .02, $z$ = 5.45, $p$<.001) than GPT-3. Whereas GPT-3 had higher utility ratings ($\beta$=-.55, SE = .06, $z$ = -8.52, $p$<.001), see Figure 5. In both groups, originality and utility were negatively correlated ($r$ = -.56 for humans and $r$ = -.61 for GPT-3).

**Do humans or GPT-3 show more flexibility in their response patterns?**

We computed flexibility scores for both humans and GPT-3 on the AUT tin can data. Humans had, on average, higher flexibility scores ($F(1, 85) = 5.53, p = .021$). However, as can be seen in Figure 6, GPT-3's flexibility scores show greater variance. GPT-3's flexibility scores were not related to temperature ($r = .04, p = .80$).

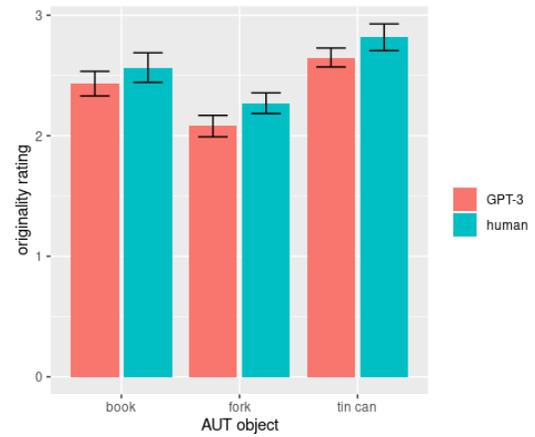

Figure 2. Human versus GPT-3 originality ratings. Human responses are rated to be more original.

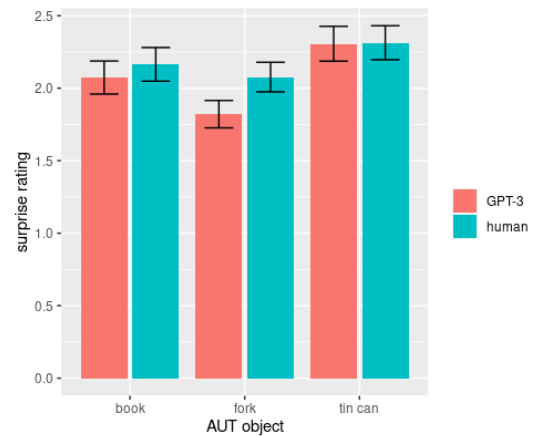

Figure 3. Human versus GPT-3 surprise ratings. Human responses are rated to be more surprising, but it's a close call.

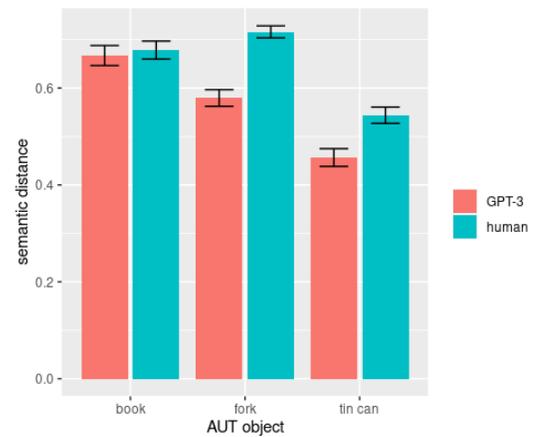

Figure 4. Human versus GPT-3 on semantic distance between response and AUT object embeddings, a proxy for creativity.

---

[1] In Figure 1, the human responses are on the left and GPT-3's responses are on the right.

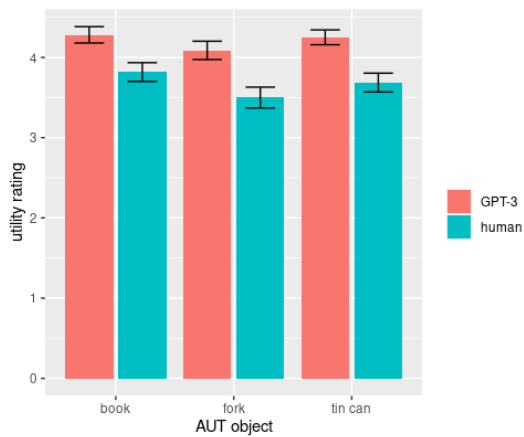

Figure 5. Human versus GPT-3 utility ratings. GPT-3 responses are rated to be more useful.

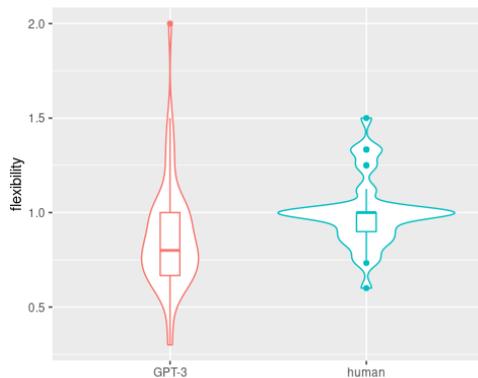

Figure 6. Human versus GPT-3 flexibility scores. Humans have a higher mean flexibility score, but GPT-3's scores show greater variance.

## Discussion

This study aimed to put GPT-3 creativity to the test using the popular Alternative Uses Test (AUT, Guilford, 1967), where participants have to come up with creative uses for an everyday object. We compared a group of psychology students' performance to different runs with GPT-3 using the criteria creativity researchers generally assess on the AUT: the originality and usefulness of responses, as well as the often discounted component, surprise. Human responses were rated higher on originality and surprise. Similarly, the semantic distance scores between the AUT object and a response, which can be considered a proxy for creativity (Beaty and Johnson, 2021), were greater for human responses. However, GPT-3's responses were rated as more useful. In both groups, the originality-utility trade-off was apparent. In general, originality weighs in more when assessing creativity (Diedrich et al., 2015), so in this case the human responses would be considered more creative.

We also compared how flexible the response sets of humans and GPT-3 were, where flexibility was computed by dividing the number of categories present in the response set by the total number of responses. So, if a response set contained five responses stemming from three categories then the flexibility score was 3/5. Humans had higher flexibility scores, but there was greater variance in GPT-3 flexibility scores. It is unclear why GPT-3's flexibility scores are more variable; it is not a function of the temperature. We leave a more thorough investigation of the flexibility of responses for future work.

The main limitation of our study is the question of whether the Alternative Uses Task, a divergent thinking task, even measures creativity (Runco, 2008; Stevenson, Baas and van der Maas, 2020). Even assuming that it only measures one aspect of creativity, we believe that comparing AI and human performance can provide us with unique insights into what creativity is and how to best measure it.

Another limitation is that our Monte Carlo experiments to determine the best combination of instructions and parameters for GPT-3 to provide optimal creative responses were not as fine-grained as we would have liked. And, on the other hand, when we administered the Remote Associates Test (Mednick, 1968) and various creative insight problems (e.g., the egg problem, Sternberg and Davidson, 1982) we received responses that seemed to have been taken verbatim from journal articles or manuals. It appears likely that many creativity tests were present in GPT-3's training data.

A final limitation concerns our human sample, not only is it small, limited to college students, but it also consisted of native Dutch speakers. So, the results of this pilot study do not necessarily generalize to most humans. Also, in order to better compare GPT-3's responses to those of our human sample we had to translate the Dutch responses to English before scoring and analyzing the data. Some creative subtleties may have been lost in translation. Furthermore, humans received only minimal instructions, whereas we optimized instructions and parameters for GPT-3. Was it a fair fight? In future work we plan to administer the AUT and other newly developed creativity tasks with optimal instructions for both humans and AI and collect data in the same language.

At this point in time, we can conclude that GPT-3's performance on the AUT is not as creative as this sample of psychology students. But, at the same time GPT-3's performance is impressive and in many cases appears human-like. A Turing test is a logical next step in this line of research. We can imagine a future in which GPT-3 and other generative LLMs responses cannot be distinguished from humans, although the creative process will be different. This is where the question arises as to the role of process in defining what is creative and what is not; we agree with Boden (2004) and Simonton (2018), that the process matters, e.g., a brute-force process is not creative, but what is? We hope that this continued line of work will provide insight into what it means to be creative, and perhaps even what it means to be human.

# References

Aalho, J., 2021. *Aum Golly – poems on humanity by an artificial intelligence.* https://aumgolly.fi/english/

Baer, J. and McKool, S.S., 2009. Assessing creativity using the consensual assessment technique. In *Handbook of research on assessment technologies, methods, and applications in higher education* (pp. 65-77). IGI Global.

Beaty, R.E. and Johnson, D.R., 2021. Automating creativity assessment with SemDis: An open platform for computing semantic distance. *Behavior research methods*, 53(2):757-780.

Bender, E.M., Gebru, T., McMillan-Major, A. and Mitchell, S., 2021, March. On the Dangers of Stochastic Parrots: Can Language Models Be Too Big?. In *Proceedings of the 2021 ACM Conference on Fairness, Accountability, and Transparency.* 610-623. https://doi.org/10.1145/3442188.3445922

Boden, M. 2004. *The creative mind: Myths and mechanisms.* Routledge.

Brown, T., Mann, B., Ryder, N., Subbiah, M., Kaplan, J.D., Dhariwal, P., Neelakantan, A., Shyam, P., Sastry, G., Askell, A. and Agarwal, S., 2020. Language models are few-shot learners. *Advances in neural information processing systems*, 33, pp.1877-1901. https://arxiv.org/abs/2005.14165

Diedrich, J., Benedek, M., Jauk, E. and Neubauer, A.C., 2015. Are creative ideas novel and useful?. *Psychology of Aesthetics, Creativity, and the Arts,* 9(1), p.35-40. https://doi.org/10.1037/a0038688

GPT-3. 2020. A robot wrote this entire article. Are you scared yet, human? *The Guardian.* https://www.theguardian.com/commentisfree/2020/sep/08/robot-wrote-this-article-gpt-3

Green, O., 2020. *Bob The Robot: Exploring the Universe - A Cozy Bedtime Story Produced by Artificial Intelligence*. Stolkholm: Olle Green.

Guilford, J.P., 1967. Creativity: Yesterday, today and tomorrow. *The Journal of Creative Behavior*, 1(1):3-14.

Hass, R.W., 2017. Semantic search during divergent thinking. *Cognition,* 166, pp.344-357. https://doi.org/10.1016/j.cognition.2017.05.039

Mednick, S.A., 1968. The remote associates test. *The Journal of Creative Behavior. 2*:213-214.

Mitchell, M., 2021. Abstraction and analogy-making in artificial intelligence. *Annals of the New York Academy of Sciences,* 1505(1):79-101. https://doi.org/10.1111/nyas.14619

Nijstad, B.A., De Dreu, C.K., Rietzschel, E.F. and Baas, M., 2010. The dual pathway to creativity model: Creative ideation as a function of flexibility and persistence. *European review of social psychology,* 21(1), pp.34-77. https://doi.org/10.1080/10463281003765323

Rietzschel, E. F., Nijstad, B. A., and Stroebe, W. 2019. Why great ideas are often overlooked. *The Oxford handbook of group creativity and innovation,* 179-197.

Runco, M.A., 2008. Commentary: Divergent thinking is not synonymous with creativity. *Psychology of Aesthetics, Creativity, and the Arts,* 2(2):93–96. https://doi.org/10.1037/1931-3896.2.2.93

Runco, M.A. and Jaeger, G.J., 2012. The standard definition of creativity. *Creativity research journal*, 24(1):92-96. https://doi.org/10.1080/10400419.2012.650092

Silvia, P.J., Martin, C. and Nusbaum, E.C., 2009. A snapshot of creativity: Evaluating a quick and simple method for assessing divergent thinking. *Thinking Skills and Creativity,* 4(2):79-85. https://doi.org/10.1016/j.tsc.2009.06.005

Simonton, D.K., 2018. Defining creativity: Don't we also need to define what is not creative?. The Journal of Creative Behavior, 52(1), pp.80-90. https://doi.org/10.1002/jocb.137

Sternberg, R.J. and Davidson, J.E., 1982. The mind of the puzzler. *Psychology Today*, 16(6):37-44.

Stevenson, C.E., Baas, M. and van der Maas, H., 2021. A minimal theory of creative ability. *Journal of Intelligence*, 9(1):9. https://doi.org/10.3390/jintelligence9010009

van der Maas, H.L., Snoek, L. and Stevenson, C.E., 2021. How much intelligence is there in artificial intelligence? A 2020 update. *Intelligence*, 87:101548. https://doi.org/10.1016/j.intell.2021.101548

# Acknowledgements

We thank Emma van Lipzig for her contribution to the data categorization. This research was partly supported by the Jacobs Foundation Fellowship 2019-2022 awarded to Claire Stevenson (2018 1288 12).

# Author Contributions

CS conceived and designed the study, collected the data, performed the analysis and wrote the original draft of the paper. IS and MB performed data coding and helped edit and review the camera ready paper. RG helped design the study and contributed to data analysis. HvdM helped conceive the study and provided supervision.